\def\year{2020}\relax
\documentclass[letterpaper]{article} 
\usepackage{aaai20}  
\usepackage{times}  
\usepackage{helvet} 
\usepackage{courier}  
\usepackage[hyphens]{url}  
\usepackage{graphicx} 
\usepackage{graphicx}  
\usepackage{bm}
\usepackage{amsmath}
\usepackage{amsthm}
\usepackage{comment}
\usepackage{algorithmic}
\usepackage{algorithm}
\usepackage{mdwmath}
\usepackage{enumerate}
\usepackage{mathrsfs}
\usepackage{graphicx}        
\usepackage{multicol}        
\usepackage{subcaption}
\usepackage{epsfig,amssymb,latexsym}
\usepackage{psfrag}
\usepackage{fancyhdr}
\usepackage{multirow}
\usepackage{color}

\newcommand{\mb}{\mathbb}
\newcommand{\mr}{\mathrm}
\newcommand{\mc}{\mathcal}

\newcounter{pro_counter}

\newtheorem{proposition}[pro_counter]{Proposition}

 \nocopyright  

\title{Lifelong Spectral Clustering}
\author{{Gan Sun}\textsuperscript{1, 3}\thanks{The corresponding author is Gan Sun.}, {Yang Cong}\textsuperscript{2}, {Qianqian Wang}\textsuperscript{3}, {Jun Li}\textsuperscript{4}, {Yun Fu}\textsuperscript{3} \\
\textsuperscript{1}{University of Chinese Academy of Sciences, China.}
\thanks{This work has been done during Gan Sun visiting Northeastern University.}\\
\textsuperscript{2}{State Key Laboratory of Robotics, Shenyang Institute of Automation, Chinese Academy of Sciences, China.}\\
\textsuperscript{3}{Northeastern University, USA.} \textsuperscript{4}{MIT, USA.}   \\
\{sungan1412, congyang81\}@gmail.com, qianqian174@foxmail.com, junl.mldl@gmail.com, yunfu@ece.neu.edu }

\begin{document}

\maketitle

\begin{abstract}
In the past decades, spectral clustering (SC) has become one of the most effective clustering algorithms. However, most previous studies focus on spectral clustering tasks with a fixed task set, which cannot incorporate with a new spectral clustering task without accessing to previously learned tasks. In this paper, we aim to explore the problem of spectral clustering in a lifelong machine learning framework, \emph{i.e.,} \underline{L}ife\underline{l}ong \underline{S}pectral \underline{C}lustering ($\mr{L^2SC}$). Its goal is to efficiently learn a model for a new spectral clustering task by selectively transferring previously accumulated experience from knowledge library. Specifically, the knowledge library of $\mr{L^2SC}$ contains two components: 1) orthogonal basis library: capturing latent cluster centers among the clusters in each pair of tasks; 2) feature embedding library: embedding the feature manifold information shared among multiple related tasks. As a new spectral clustering task arrives, $\mr{L^2SC}$ firstly transfers knowledge from both basis library and feature library to obtain encoding matrix, and further redefines the library base over time to maximize performance across all the clustering tasks. Meanwhile, a general online update formulation is derived to alternatively update the basis library and feature library. Finally, the empirical experiments on several real-world benchmark datasets demonstrate that our $\mr{L^2SC}$ model can effectively improve the clustering performance when comparing with other state-of-the-art spectral clustering algorithms.
\end{abstract}

\section{Introduction}
Spectral clustering algorithms \cite{ng2002spectral,shi2000normalized} discover the corresponding embedding of data via utilizing manifold information embedded in the sample distribution, which has shown the state-of-the-art performance in many applications \cite{li2015superpixel,zhao2017multi,Seg_Lichen_TIP18}. In addition to single spectral clustering task scenario, \cite{yang2015multitask} proposes a multi-task spectral clustering model, and aims to perform multiple clustering tasks and make them reinforce each other. However, most recently-proposed models \cite{zhang2018multi,pang2018spectral,kang2018unified} focus on clustering tasks with a fixed task set. When applied into a new task environment or incorporated into a new spectral clustering task, these models have to repeatedly access to previous clustering tasks, which can result in high energy consumption in real applications, \emph{e.g.,} in mobile applications. In this paper, our work explores how to adopt the spectral clustering scenario into the setting of lifelong machine learning.

\begin{figure}[t]
\centering
\includegraphics[width=.95\columnwidth]{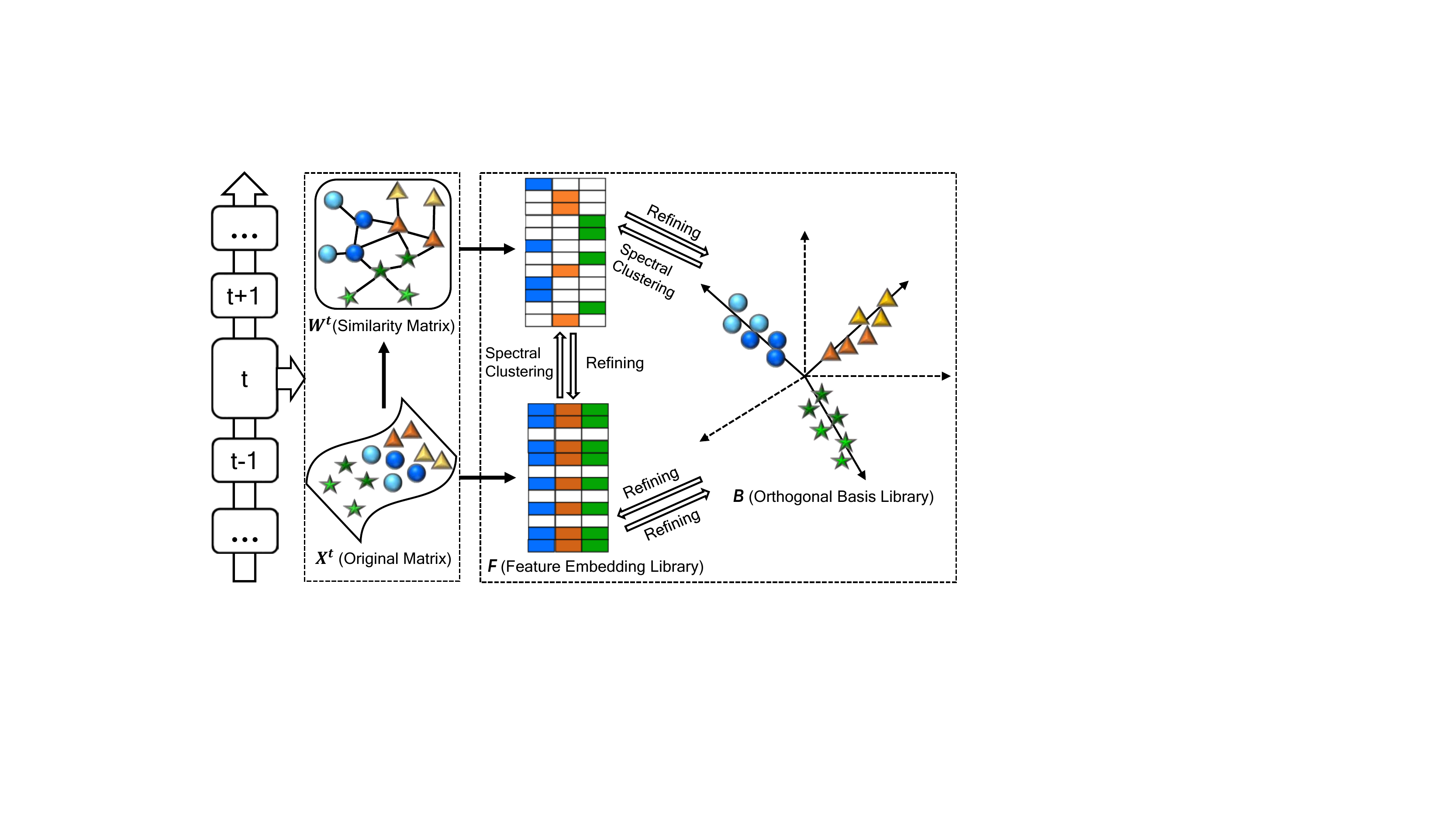}
\caption{The demonstration of our lifelong spectral clustering model, where different shapes are from different clusters. When a new clustering task $X^t$ is coming, the knowledge is iteratively transferred from orthogonal basis library $B$ and feature embedding library $F$ to encode the new task.}
\label{fig:lifelongspectralclustering}
\end{figure}


For the lifelong machine learning, recent works \cite{ruvolo2013ella,isele2016using,xu2018lifelong,sun2018robust,sun2019representative} have explored the methods of accumulating the single task over time. Generally, lifelong learning utilizes knowledge from previously learned tasks to improve the performance on new tasks, and accumulates a knowledge library over time. Although these models have been successfully adopted into supervised learning \cite{chen2018lifelong,sun2018active} and reinforcement learning \cite{ammar2014online,isele2018selective}, its application in spectral clustering, one of the most classical research problems in machine learning community, is still sparse. \emph{Take the news clustering tasks as an example, the semantic meaning of \emph{Artificial Intelligence} and \emph{NBA} are very dissimilar in the newspaper of year 2010, and should be divided into different clusters. The clustering task of year 2010 can thus contribute to the clustering task of year 2020 in a never-ending perspective, since the correlation information between \emph{Artificial Intelligence} and \emph{NBA} of year 2020 is similar with that in year 2010.}

Inspired by the above scenario, this paper aims to establish a lifelong learning system with spectral clustering tasks, \emph{i.e.,} lifelong spectral clustering. Generally, the main challenges among multiple consecutive clustering tasks are as follows: 1) \textbf{Cluster Space Correlation:} the latent cluster space should be consistent among multiple clustering tasks. For example, for the news clustering task, the cluster centers in year 2010 can be \{Business, Technology, Science, etc\}, while the cluster centers in year 2020 are similar to that in year 2010; 2) \textbf{Feature Embedding Correlation:} another correlation among different clustering tasks is feature correlation. For example, in consecutive news cluster tasks, the semantic meaning of \emph{Artificial Intelligence} are very similar in year 2010 and year 2020. Thus, the feature embedding of \emph{Artificial Intelligence} should be same for these two tasks.




To tackle the challenges above, as shown in Figure~\ref{fig:lifelongspectralclustering}, we propose a \underline{L}ife\underline{l}ong \underline{S}pectral \underline{C}lustering (\emph{i.e.,} $\mr{L^2SC}$) model by integrating cluster space and feature embedding correlations, which can achieve never-ending knowledge transfer between previous clustering tasks and later ones. To achieve this, we present two knowledge libraries to preserve the common information among multiple clustering tasks, \emph{i.e.,} orthogonal basis and feature embedding libraries. Specifically, 1) orthogonal basis library contains a set of latent cluster centers, \emph{i.e.,} each sample of cluster tasks can be effectively assigned to multiple clusters with different weights; 2) feature embedding library can be modeled by introducing bipartite graph co-clustering, which can not only discover the shared manifold information among cluster tasks, but also maintain the data manifold information of each individual task. When a new spectral clustering task is coming, $\mr{L^2SC}$ can firstly encode the new task via transferring the knowledge of both orthogonal basis library and feature embedding library to encode the new task. Accordingly, these two libraries can be refined over time to keep on improving across all clustering tasks. For model optimisation, we derive a general lifelong learning formulation, and further optimize this optimization problem via applying an alternating direction strategy. Finally, we evaluate our proposed model against several spectral clustering algorithms and even multi-task clustering models on several datasets. The experimental results strongly support our proposed $\mr{L^2SC}$ model.

\indent The novelties of our proposed $\mr{L^2SC}$ model include:
\begin{itemize}
  \item To our best knowledge, this work is the first attempt to study the problem of spectral clustering in the lifelong learning setting, \emph{i.e.,} Lifelong Spectral Clustering ($\mr{L^2SC}$), which can adopt previously accumulated experience to incorporate new cluster tasks, and improve the clustering performance accordingly.
  \item We present two common knowledge libraries: orthogonal basis library and feature embedding libray, which can simultaneously preserve the latent clustering centers and capture the feature correlations among different clustering tasks, respectively.
  \item We propose an alternating direction optimization algorithm to optimize the proposed $\mr{L^2SC}$ model efficiently, which can incorporate fresh knowledge gradually from online dictionary learning perspective. Various experiments show the superiorities of our proposed model in terms of effectiveness and efficiency.
\end{itemize}

\section{Related Work}\label{sec:related work}
In this section, we briefly provide a review on two topics: \textbf{Multi-task Clustering} and \textbf{Lifelong Learning}.




For the \textbf{Multi-task Clustering} \cite{zhang2018multi}, the learning paradigm is to combine multi-task learning~\cite{sun2017joint} with unsupervised learning, and the key issue is how to transfer useful knowledge among different clustering tasks to improve the performance. Based on this assumption, recently-proposed methods \cite{zhang2017multi,huy2013feature} achieve knowledge transfer for clustering via using some sample from other tasks to form better distance metrics or $K$-nn graphs. However, these methods ignore employing the task relationships in the knowledge transfer process. To preserve task relationships, multi-task Bregman clustering (MBC) \cite{zhang2011multitask} captures the task relationships by alternatively update clusters among different tasks.  For the spectral clustering based multi-task clustering, multi-task spectral clustering (MTSC) \cite{yang2015multitask} take the first attempt to extend spectral clustering into multi-task learning. By using the inter-task and intra-task correlations, a $\ell_{2,p}$-norm regularizer is adopted in MTSC to constrain the coherence of all the tasks based on the assumption that a low-dimensional representation is shared by related tasks. Then a mapping function is learned to predict cluster labels for each individual task.


\indent For the \textbf{Lifelong Learning}, the early works on this topic focus on transferring the selective information from task cluster to the new tasks \cite{thrun1996discovering,sun2018lifelong}, or transferring invariance knowledge in neural networks \cite{thrun2012explanation}. In contrast, an efficient lifelong learning algorithm (ELLA) \cite{ruvolo2013ella} is developed for online learning multiple tasks in the setting of lifelong learning. By assuming that models of all related tasks share a common basis, each new task can be obtained by transferring knowledge from the basis. Furthermore, \cite{ammar2014online} extends this idea into learn decision making tasks consecutively, and achieves dramatically accelerate learning on a variety of dynamical systems; \cite{isele2016using} proposes a coupled dictionary to incorporate task descriptors into lifelong learning, which can enable performing zero-shot transfer learning. Since observed tasks in lifelong learning system may not compose an \emph{i.i.d} samples, learning an inductive bias in form of a transfer procedure is proposed in \cite{pentina2015lifelong}. Different from traditional learning models \cite{rannen2017encoder}, \cite{li2016learning} proposes a learning without forgetting method for convolutional neural network, which can train the network only using the data of the new task, and retain performance on original tasks via knowledge distillation \cite{hinton2015distilling}, and train the network using only the data of the new task. Among the discussion above, there is no works concerning lifelong learning in the spectral clustering setting, and our current work represents the first work to achieve lifelong spectral clustering.





\section{Lifelong Spectral Clustering ($\mr{L^2SC}$)}\label{sec:formulation}
This section introduces our proposed lifelong spectral clustering ($\mr{L^2SC}$) learning model. Firstly, we briefly review a general spectral clustering formulation for single spectral clustering task. Our $\mr{L^2SC}$ model for lifelong spectral clustering task problem is then given.



\subsection{Revisit Spectral Clustering Algorithm}
This subsection reviews a general spectral clustering algorithm with normalized cut. Given an undirected similarity graph $G^t=\{X^t,W^t\}$ with a vertex set $X^t\in \mb{R}^{d\times n_t}$ and an corresponding affinity matrix $W^t\in \mb{R}^{n_t\times n_t}$ for the clustering task $t$, where $d$ is the number of the features, $n_t$ is the total number of data samples for the task $t$, each element $w_{ij}^t$ in symmetric matrix $W^t$ denotes the similarity between a pair of vertices $(x_i^t,x_j^t)$. The common choice for matrix $W^t$ can be defined as follows:
\begin{equation*}
w_{ij}^t=\begin{cases}
\mr{exp}\Big(\!\!-\!\frac{\left\|x_i^t-x_j^t\right\|^2}{2\sigma^2}\!\Big), \; \mr{if}\; x_i^t \!\in \! \mc{N}(x_j^t)\, \mr{or}\, x_j^t \!\in \! \mc{N}(x_i^t) \\
0, \quad\quad\quad\quad\quad\quad\quad\quad\quad\quad\quad  \mr{otherwise}, \\
\end{cases}
\end{equation*}
where $\mc{N}(\cdot)$ is the function for searching $K$-nearest neighbors, and $\sigma$ controls the spread of the neighbors. After applying the normalized Laplacian:
\begin{equation}\label{eq:spectralcluster_k}
W_N^t=(D^t)^{-\frac{1}{2}}L^t(D^t)^{-\frac{1}{2}}=I-(D^t)^{-\frac{1}{2}}W^t(D^t)^{-\frac{1}{2}},
\end{equation}
where $D^t$ is a diagonal matrix with the diagonal elements as $D_{ii}^t=\sum_{j}w_{ij}^t\, (\forall i)$. The final formulation of spectral clustering turns out to be the well-known normalized cut \cite{shi2000normalized}, and can be expressed as:
\begin{equation}\label{eq:spectralcluster_single}
  \max_{F^t}\; \mr{tr}\big((F^t)^{\top}K^tF^t\big),\quad s.t., (F^t)^{\top}F^t=I_k,
  \end{equation}
where $K^t=(D^t)^{-\frac{1}{2}}W^t(D^t)^{-\frac{1}{2}}$, the optimal cluster assignment matrix $F^t$ can be achieved via eigenvalue decomposition of matrix $K^t$. Based on the relaxed continuous solution, then the final discrete solution of $F^t$ can be obtained by spectral rotation or $K$-means, \emph{e.g.,} the $j$-th element of $f_i^t$ is $1$, if the sample $x_i^t$ is assigned to the $j$-th cluster; $0$, otherwise.




\subsection{Problem Statement}
Given a set of $m$ unsupervised clustering tasks $\mc{T}^1,\ldots,\mc{T}^m$, where each individual clustering task $\mc{T}^t$ has a set of $n_t$ training data samples $X^t\in \mb{R}^{d\times n_t}$, and the dimensionality of feature space is $d$. The original intention of multi-task spectral clustering method \cite{yang2015multitask} is to uncover the correlations among all the clustering tasks, and predict the cluster assignment matrices $\{F^t\}_{t=1}^m$ for each clustering task. However, learning incremental spectral clustering tasks without accessing to the previously-adopted clustering data is not considered in traditional single or multi-task spectral clustering models. In the setting of spectral clustering, a lifelong spectral clustering system encounters a series of spectral clustering tasks $\mc{T}^1,\ldots,\mc{T}^m$, where each task $\mc{T}^t$ is defined in Eq.~\eqref{eq:spectralcluster_single}, and intends to obtain new cluster assignment matrix $F^t \in \mb{R}^{n_t\times k}$ for the task $t$. For convenience, this paper assume that the learner in this lifelong machine learning system do not know any information about clustering tasks, \emph{e.g.,} the task distributions, the total number of spectral clustering tasks $m$, etc. When lifelong spectral clustering system receives a batch of data for some spectral clustering task $t$ (either a new spectral clustering task or previously learning task $t$) in each period, this system should obtain cluster assignment matrix of samples of encountered tasks. The goal is to obtain corresponding task assignment matrices $F^1, \ldots,F^m$ such that: \textbf{1)} Clustering Performance: each obtained assignment matrix $F^t$ should preserve the data configuration of the $t$-th task, and partition the new clustering task more accurate; \textbf{2)} Computational Speed: in each clustering period, obtaining each $F^t$ should be faster than that among traditional multi-task spectral clustering methods; \textbf{3)} Lifelong Learning: new $F^t$'s can be arbitrarily and efficiently added when the lifelong clustering system faces with new unsupervised spectral clustering tasks.


\subsection{The Proposed $\mr{L^2SC}$ Model}
In this section, we introduce how to model the lifelong learning property and cross-task correlations simultaneously. Basically, there are two challenges in the $\mr{L^2SC}$ model:

\textbf{1) Orthogonal Basis Library:} in order to achieve lifelong learning, one of the major component is how to store the previously accumulated experiences, \emph{i.e.,} knowledge library. To tackle this issue, inspired by \cite{han2015unsupervised} which employs the orthogonal basis clustering to uncover the latent cluster centers, each assignment matrix $F^t$ can be decomposed into two submatrices, \emph{i.e.,} a basis matrix $B\in \mb{R}^{k\times k}$ called orthogonal basis library, and a cluster encoding matrix $E^t\in\mb{R}^{n_t \times k}$, as $F^t=E^tB$. Then the multi-task spectral clustering formulation can be expressed as:
\begin{equation}
\begin{aligned}\label{eq:final_problem1}
\max_{\atop \{E^t\}_{t=1}^m} &   \frac{1}{m}\sum_{t=1}^m\! \mr{tr}\big(\big(E^tB\big)^{\top}\!K^tE^tB\big), \\
s.t., & B^{\top}B=I_k, (E^t)^{\top}E^t\!=\!I_k,  \forall t=1,\ldots,m,
\end{aligned}
\end{equation}
  where the orthogonal constraint of matrix $B$ encourages each column of $B$ to be independent, and $K^t$ is defined in the Eq.~\eqref{eq:spectralcluster_k}. Therefore, the orthogonal basis library $B$ can be used to refine the latent cluster centers and further obtain an excellent cluster separation.

\textbf{2) Feature Embedding Library:} even though the latent cluster centers can be captured gradually in Eq.~\eqref{eq:final_problem1}, it does not consider the common feature embedding transfer across multiple spectral clustering tasks. Motivated by \cite{jiang2012transfer} which adopts graph based co-clustering to control and achieve the knowledge transfer between two tasks, we propose to link each pair of clustering tasks together such that one embedding obtained in one task can facilitate the discover of the embedding in another task. We thus define an invariant feature embedding library $L\in \mb{R}^{d\times k}$ with group sparse constraint, and give the graph co-clustering term as:
\begin{equation}\label{eq:final_problem2}
  \max_{L} \frac{1}{m}\!\sum_{t=1}^m \mr{tr} (L^{\top}\hat{X}^tE^tB)\!+\!\mu\!\left\|L\right\|_{2,1},  s.t., L^{\top}L=I_k,
 \end{equation}
and $\hat{X}^t$ for the $t$-th task is defined as:
\begin{equation}\label{eq:define_X}
 \hat{X}^t =(D_1^t)^{-\frac{1}{2}}X^t(D_2^t)^{-\frac{1}{2}},
\end{equation}
 where $D_1^t=\mr{diag}(X^t\bm{1})$, and $D_2^t=\mr{diag}\big((X^t)^{\top}\bm{1}\big)$. Intuitively, with this sharing embedding library $L$, multiple spectral clustering tasks can transfer embedding knowledge with each other in a perspective of common feature learning \cite{Argyriou:2008}.

Given the same graph construction method and training data for each spectral clustering task, we solve the optimal cluster assignment matrix $\{F^t\}_{t=1}^m$ while encouraging each clustering task to share common knowledge in libraries $B$ and $L$. By combining these two goals in Eq.~\eqref{eq:final_problem1} and Eq.~\eqref{eq:final_problem2}, then lifelong spectral clustering model can be expressed as the following objective function:
\begin{equation}
  \begin{aligned}\label{eq:lsc_finalproblem}
   \max_{B,L,\{E^t\}_{t=1}^m} & \; \frac{1}{m}\sum_{t=1}^m \Big\{\mr{tr}\big(\big(E^tB\big)^{\top}K^tE^tB\big)\\
    &+\lambda_t \mr{tr}(L^T\hat{X}^tE^tB)\Big\}+\mu\left\|L\right\|_{2,1},\\
   s.t. ,\;   B^{\top}&B=I_k, L^{\top}L=I_k, (E^t)^{\top}E^t=I_k,
  \end{aligned}
 \end{equation}
where $\lambda_t$'s are the trade-off between the each spectral clustering task with the co-clustering objective. If $\lambda_t$'s are set as $0$, this model can reduce to the multi-task spectral clustering model with common cluster centers.

\begin{algorithm}[t]
\caption{ Lifelong Spectral Clustering ($\mr{L^2SC}$) Model }
\begin{algorithmic}[1]
\STATE \textbf{Input:} Spectral clustering tasks: $X^1,\ldots, X^m$, Library: $B\leftarrow \bm{0}_{k\times k}$, $L\leftarrow \bm{0}_{d\times k}$, $\mu\geq 0, \lambda_t \geq0,\forall t=1,\ldots,m$, Statistical records: $M_0 \leftarrow \bm{0}_{k\times k}$, $C_0\leftarrow \bm{0}_{d\times k}$;
\WHILE {Receive clustering task data}
\STATE  New $t$-th task: $(X^t,t)$;
\STATE  Construct matrices $\{K^t, \hat{X}^t\}$;
\WHILE  {Not Converge}
\STATE  Update $E^t$ via Eq.~\eqref{eq:optimize_E_t};
\STATE  Update $B$ via Eq.~\eqref{eq:optimize_B};
\STATE  Update $L$ via Eq.~\eqref{eq:optimize_L};
\STATE  Update $\Theta$ via $\Theta_{ii}=\frac{1}{2\left\|l_i\right\|_2}, (\forall i=1,\ldots,d)$;
\ENDWHILE
\STATE  Compute cluster assignment matrices via $\{E^tB\}_{t=1}^m$;
\STATE  Compute final indicator matrices via $K$-means;
\ENDWHILE
\end{algorithmic}
\end{algorithm}

\subsection{Model Optimization}
This section shows how to optimize our proposed $\mr{L^2SC}$ model. Normally, standard alternating direction strategy using all the learned tasks is inefficient to this lifelong learning model in Eq.~\eqref{eq:lsc_finalproblem}. Our goal in this paper is to build an lifelong clustering algorithm that both CPU time and memory space have lower computational cost than offline manner. When a new spectral clustering task $m$ arrives, the basic ideas for optimizing Eq.~\eqref{eq:lsc_finalproblem} is: both $L$, $B$ and $E^m$ should be updated without accessing to the previously learned tasks, \emph{e.g.,} the previous data in matrices $\{K^{t},\hat{X}^t\}_{t=1}^{m-1}$. In the following, we briefly introduce the proposed update rules, and provide the convergence analysis in the experiment.

\subsubsection{Updating $E^m$ with fixed $L$ and $B$:}
With the fixed $L$ and $B$, the problem for solving encoding matrix $E^m$ can be expressed as:
\begin{equation}\label{eq:optimize_E_t}
  \max_{(E^m)^{\top}E^m=I_k} \!\!\!\mr{tr}\big(\big(E^mB\big)^{\top}\!\!K^mE^mB\big)\! +\!\lambda_m  \mr{tr}\big(L^{\top}\!\hat{X}^mE^mB\big).
\end{equation}
With the orthonormality constraint, $E^m$ can be updated in the setting of Stiefel manifold \cite{manton2002optimization}, which is defined by the following Proposition.
\begin{proposition}
Let $X\in \mb{R}^{n\times k}$ be a rank $p$ matrix, where the singular value decomposition (\emph{i.e.,} SVD) of $X$ is $U\Sigma V^{\top}$. The projection of matrix $X$ on Stiefel manifold is defined as:
\begin{equation}
\pi(X)=\mathop{\arg\min}_{Q^{\top}Q=I}\; \left\|X-Q\right\|_F^2.
\end{equation}
The projection could be calculated as: $\pi(X)=UI_{n,k}V^{\top}$.
\end{proposition}
Therefore, we can update $E^m$ by moving it in the direction of increasing the value of the objective function, and the update operator can be given as:
\begin{equation}\label{eq:solution_E_t}
  E^m=\pi(E^m+\eta_m \nabla f(E^m)),
\end{equation}
where $\eta_m$ is the step size, $f(E^m)$ is the objective function of Eq.~\eqref{eq:optimize_E_t}, and $\nabla f(E^m)$ can be defined as $2(K^m)^{\top}E^mBB^{\top}+\lambda_m(\hat{X}^m)^{\top}LB^{\top}$. To guarantee the convergence of the optimization problem in Eq.~\eqref{eq:optimize_E_t}, we provide a convergence analysis at the experiment section.


\subsubsection{Updating $B$ with fixed $L$ and $\{E^t\}_{t=1}^m$:}
With the obtained encoding matrix $E^m$ for the new coming $m$-th task, the optimization problem for variable $B$ can be:

\begin{equation}\label{eq:optimize_B}
  \hspace{-3pt}\max_{B^{\top}B=I_k}\!\! \frac{1}{m}\sum_{t=1}^{m}\!\mr{tr}\big(B^{\top}\big(E^t\big)^{\top}\!K^tE^tB\big)\!+\!\lambda_t  \mr{tr}(L^{\top}\hat{X}^tE^tB).
\end{equation}
Based on the orthonormality constraint $B^{\top}B=I_k$, we can rewrite Eq.~\eqref{eq:optimize_B} as follows:

\begin{equation}
\begin{aligned}
  & \! \max_{B^{\top}B=I_k} \! \! \frac{1}{m}\sum_{t=1}^{m}\mr{tr}(B^{\top}\big(\big(E^t\big)^{\top}K^tE^t+\lambda_tBL^{\top}\hat{X}^tE^t\big)B), \\
  \Leftrightarrow  & \!\!\!\! \max_{B^{\top}B=I_k} \! \!\! \mr{tr}(B^{\top}\!\big(\frac{1}{m}\!\!\sum_{t=1}^{m}\!(E^t)^{\top}\!K^tE^t \!+ \! \frac{1}{m}\!\!\sum_{t=1}^{m}\lambda_tBL^{\top}\!\hat{X}^tE^t\big)B)
\end{aligned}
\end{equation}
To better store the previous knowledge of learned clustering tasks, we then introduce two statistical variables:
\begin{equation}\label{eq:statistical_records}
  M_m\!=\!M_{m-1}+(E^m)^{\top}K^mE^m, C_m\!=\!C_{m-1}+\lambda_m\hat{X}^mE^m,
\end{equation}
 where $M_{m-1}=\sum_{t=1}^{m-1}(E^t)^{\top}K^tE^t$, and $C_{m-1}=\sum_{t=1}^{m-1}\lambda_t\hat{X}^tE^t$. Therefore, knowledge of new task is $(E^m)^{\top}K^mE^m$ and $\hat{X}^mE^m$. With $B$ as a warm start, so:

 \begin{equation}\label{eq:solution_B}
    B=\mathop{\arg\max}_{B^{\top}B=I_k}\; \mr{tr}\big(B^{\top}\big(M_m/m+BL^{\top}C_m/m\big)B\big).
  \end{equation}
It is well-known that the solution of $B$ can be relaxedly obtained by the eigen-decomposition of $\big(M_m/m+BL^{\top}C_m/m\big)$. Notice that even though the input parameter of Eq.~\eqref{eq:solution_B} contains $B$, the above solution is also effective since the proposed algorithm converges very quickly in the online manner.


\subsubsection{Updating $L$ with fixed $B$ and $\{E^t\}_{t=1}^m$:}
With the obtained center library $B$ and encoding matrix $E^m$ for the new coming $m$-th task, the optimization problem for variable $L$ can be denoted as:
\begin{equation}
\begin{aligned}\label{eq:optimize_L}
\max_{L^{\top}L=I_k} & \; \frac{1}{m}\sum_{t=1}^m \lambda_t  \mr{tr}\big(L^{\top}\hat{X}^tE^tB\big)+\mu\left\|L\right\|_{2,1},
\end{aligned}
\end{equation}
and the equivalent optimization problem can be formulated as following equations:
\begin{equation}
\begin{aligned}\label{eq:solution_L1}
  &\min_{L^{\top}L=I_k } -\mr{tr}\big(L^{\top} (\frac{1}{m}\sum_{t=1}^m\lambda_t \hat{X}^tE^t)B+\mu \Theta L)\big),\\
  \Leftrightarrow &\min_{L^{\top}L=I_k } \left\|L-\big((\frac{1}{m}\sum_{t=1}^m\lambda_t \hat{X}^tE^t)B+\mu \Theta^{-1} L\big)\right\|_F^2,\\
 \Leftrightarrow &\min_{L^{\top}L=I_k } \left\|L-\big(C_mB+\mu \Theta^{-1} L\big)\right\|_F^2,
\end{aligned}
\end{equation}
which is also definition of projection of $(C_mB+\mu \Theta L)$ on the Stiefel manifold. Further, $\Theta$ denotes a diagonal matrix with each diagonal element as: $\Theta_{ii}=\frac{1}{2\left\|l_i\right\|_2}$ \cite{nie2010efficient}, where $l_i$ is the $i$-th row of $L$.


Finally, the cluster assignment matrices for all learned tasks can be computed via $\{E^tB\}_{t=1}^m$, and final indicator matrices are obtained using $K$-means.   The whole optimization procedure is summarized in \textbf{Algorithm 1}.

\begin{table*}[t]
\caption{Comparison results in terms of 3 different metrics (mean $\pm$ standard deviation) on WebKB4 dataset.}
\centering
\scalebox{0.762}{
\begin{tabular}{|cc||ccc||ccccc||c|}
\hline
 {}&Metrics&stSC & uSC & OnestepSC& MBC& SMBC &SMKC & MTSC  & MTCMRL &Ours \\
 \hline\hline
  \multirow {3}{*}{
  \centering Task1}
   & Purity($\%$) &62.66$\pm$0.00  & 59.78$\pm$0.31 &66.89$\pm$0.63  &63.95$\pm$4.07  &64.62$\pm$4.05      & 60.59$\pm$3.70 &65.92$\pm$0.68    &74.40$\pm$1.16  &\textbf{80.00$\pm$1.25} \\

   &NMI($\%$)& 13.95$\pm$0.00  & 13.15$\pm$1.68 & 14.56$\pm$3.44     &26.44$\pm$3.73       &25.53$\pm$2.74      & 14.14$\pm$4.38 &25.73$\pm$0.98 & 38.71$\pm$1.47   &\textbf{49.07$\pm$1.41} \\

   &RI($\%$)& 59.89$\pm$0.00  & 58.83$\pm$0.04 &64.76$\pm$1.06  &61.64$\pm$3.58  &62.58$\pm$2.65      & 59.45$\pm$1.62  &62.85$\pm$0.76     &73.47$\pm$0.64 &\textbf{79.05$\pm$3.67} \\ \hline
   \multirow {3}{*}{
  \centering Task2}
    &  Purity($\%$) & 62.00$\pm$0.00  & 67.00 $\pm$0.28 & 68.40$\pm$0.02  &68.12$\pm$1.81  &68.06$\pm$0.92  &60.73$\pm$2.56 &69.00$\pm$0.84 & 72.08$\pm$2.19 &\textbf{74.40$\pm$1.13}  \\

    & NMI($\%$) &  16.72$\pm$0.00  & 20.28 $\pm$1.81 & 20.56$\pm$2.39 &27.22$\pm$3.92  &27.02$\pm$3.61  &13.58$\pm$3.52 & 26.57$\pm$1.63 & 33.42$\pm$3.25 &\textbf{41.89$\pm$1.49}  \\

    & RI($\%$) & 57.12$\pm$0.00  & 60.38$\pm$2.06  & 64.81$\pm$1.52  &68.04$\pm$2.46  &68.32$\pm$3.29 &58.31$\pm$ 1.19 & 66.57$\pm$0.85 & 69.94$\pm$1.72 &\textbf{74.79$\pm$0.13}  \\   \hline
  \multirow {3}{*}{
  \centering Task3}
   & Purity($\%$) &  69.21$\pm$0.27 & 59.80$\pm$0.27& 69.80$\pm$0.55  &64.86$\pm$5.36 & 68.04$\pm$2.28 &66.01$\pm$4.13  & 68.23$\pm$0.55 & \textbf{76.47$\pm$3.15} & 74.12$\pm$1.10 \\

   & NMI($\%$) &  29.24$\pm$0.30 & 15.60$\pm$2.42 & 22.55$\pm$2.36  &26.50$\pm$3.97 & 28.32$\pm$3.86 &22.09$\pm$5.95  & 29.33$\pm$0.99 &40.97$\pm$5.26 &  \textbf{44.69$\pm$3.68} \\

   & RI($\%$) &  66.57$\pm$0.19 & 61.84$\pm$0.60  & 66.16$\pm$0.22   &65.86$\pm$4.09 &67.34$\pm$3.23 &65.02$\pm$2.41  & 65.56$\pm$0.87 &76.34$\pm$4.85 &  \textbf{78.53$\pm$1.97} \\ \hline
 \multirow {3}{*}{
  \centering Task4}
   & Purity($\%$) & 69.61$\pm$0.00 &70.42$\pm$0.23  & 71.31$\pm$0.92 &72.18$\pm$4.17 & 71.21$\pm$4.08 &69.82$\pm$2.58  &69.93$\pm$0.46 &78.23$\pm$2.68 &\textbf{80.06$\pm$0.18} \\

   & NMI($\%$) & 33.75$\pm$0.00  &33.15$\pm$0.49  & 36.84$\pm$0.59 &39.97$\pm$5.24 &39.53$\pm$2.74 &30.31$\pm$4.17   & 45.64$\pm$0.66 &49.23$\pm$2.17 &\textbf{49.26$\pm$0.79} \\

   & RI($\%$) & 66.93$\pm$0.00  &67.50$\pm$0.54   &68.69$\pm$0.94  &70.27$\pm$3.59 &70.29$\pm$2.65 &67.62$\pm$1.85  &60.72$\pm$1.15 &\textbf{79.01$\pm$1.54} &77.94$\pm$0.97 \\  \hline\hline
 \multirow {3}{*}{
  \centering }
   & Avg.Purity($\%$) & 65.87$\pm$0.07 &64.25$\pm$0.27  &69.10$\pm$0.53 &67.28$\pm$3.85 & 67.98$\pm$2.83 &64.29$\pm$3.24  &68.27$\pm$0.64 &75.19$\pm$2.25 &\textbf{77.14$\pm$0.92} \\

   &Avg.NMI($\%$) & 23.42$\pm$0.07  &20.55$\pm$1.60  &23.63$\pm$2.19 &30.03$\pm$4.22 &30.10$\pm$4.05 &20.03$\pm$4.50 &31.82$\pm$1.07 &40.58$\pm$3.04 &\textbf{46.26$\pm$1.84} \\

   &Avg.RI($\%$) & 62.63$\pm$0.05  &62.14$\pm$0.81   &66.11$\pm$0.94 &66.45$\pm$3.43 &70.29$\pm$2.65 &62.60$\pm$1.76  &63.93$\pm$0.91 &74.69$\pm$2.19 &\textbf{77.58$\pm$1.68} \\  \hline

 \end{tabular}
}
\label{table:WebKB4}
\end{table*}

\begin{table*}[t]
\caption{Comparison results in terms of 3 different metrics (mean $\pm$ standard deviation) on Reuters dataset.}
\centering
\scalebox{0.77}{
\begin{tabular}{|cc||ccc||ccccc||c|}
\hline
 {}&Metrics&stSC & uSC & OnestepSC& MBC& SMBC &SMKC & MTSC  & MTCMRL &Ours \\
 \hline\hline
  \multirow {3}{*}{
  \centering Task1}
   & Purity($\%$) &95.63$\pm$0.00  & 85.44$\pm$0.00 &94.66$\pm$0.00  &73.30$\pm$9.27  &89.90$\pm$1.40 & 95.75$\pm$0.72 &97.57$\pm$0.00    &97.57$\pm$0.00   & \textbf{98.06$\pm$0.00} \\

   &NMI($\%$)& 82.72$\pm$0.00  & 60.54$\pm$0.00 & 75.89$\pm$1.52   &61.39$\pm$2.32   &77.92$\pm$3.31   & 84.17$\pm$2.05  &89.49$\pm$0.00 & 89.49$\pm$0.00  &\textbf{91.19$\pm$0.00} \\

   &RI($\%$)& 94.64$\pm$0.00  & 82.22$\pm$0.00 & 91.44$\pm$1.06  &73.83$\pm$7.26  &88.35$\pm$1.77  & 94.35$\pm$0.88  &96.83$\pm$0.00  &96.83$\pm$0.00 & \textbf{97.43$\pm$0.00} \\ \hline
   \multirow {3}{*}{
  \centering Task2}
    &  Purity($\%$) & 84.62$\pm$0.00  & 70.00$\pm$0.00 & 86.92$\pm$0.00  &70.19$\pm$0.73  &92.88$\pm$0.38  &90.96$\pm$1.15 &96.15$\pm$0.54   & 97.31$\pm$0.54&\textbf{98.23$\pm$0.09}  \\

    & NMI($\%$)     & 62.91$\pm$0.00  & 53.17$\pm$0.00 & 64.45$\pm$0.00 & 53.43$\pm$7.81  &79.54$\pm$1.27  &75.76$\pm$2.65 &84.89$\pm$1.62   & 88.93$\pm$2.46 &\textbf{91.70$\pm$1.01}  \\

    & RI($\%$)      & 80.83$\pm$0.00  & 75.95$\pm$0.00  & 82.52$\pm$0.00  &71.77$\pm$1.08 &90.44$\pm$0.44 &88.12$\pm$1.35 &95.07$\pm$0.55   & 96.41$\pm$0.77 &\textbf{98.11$\pm$0.05}  \\   \hline
     \multirow {3}{*}{
  \centering Task3}
   & Purity($\%$) & 75.26$\pm$0.00 & 82.63$\pm$0.00  &76.05$\pm$1.86 &72.36$\pm$9.78 &75.24$\pm$2.98  &76.50$\pm$2.07  &90.79$\pm$0.37 &94.21$\pm$0.00 &\textbf{95.26$\pm$0.74} \\

   & NMI($\%$)    & 54.00$\pm$0.00  &59.85$\pm$0.00  & 61.74$\pm$1.44 &46.35$\pm$6.70 &54.11$\pm$5.41  &52.72$\pm$2.79 &73.37$\pm$0.66 &\textbf{79.45$\pm$0.00} &78.62$\pm$0.47 \\

   & RI($\%$)     & 70.14$\pm$0.00  &78.01$\pm$0.00   &74.64$\pm$1.54 &74.34$\pm$3.64 &70.01$\pm$4.33 &72.73$\pm$2.89  &88.33$\pm$0.49 &\textbf{93.13$\pm$0.00} &93.07$\pm$0.51 \\  \hline \hline
 \multirow {3}{*}{
  \centering }
   & Avg.Purity($\%$) &85.17$\pm$0.00 &79.36$\pm$0.00 &85.88$\pm$0.62  &71.95$\pm$6.59 &86.01$\pm$1.59 &87.74$\pm$1.32 &94.96$\pm$0.46  &96.36$\pm$0.18 &\textbf{97.18$\pm$0.74} \\

   &Avg.NMI($\%$)     &66.54$\pm$0.00 &79.36$\pm$0.18 &67.35$\pm$0.99 &53.72$\pm$5.61 &70.52$\pm$3.33    &70.88$\pm$2.50 &83.63$\pm$1.14  &85.96$\pm$0.82 &\textbf{87.71$\pm$0.47} \\

   &Avg.RI($\%$)      &81.87$\pm$0.00 &78.73$\pm$0.90 &82.87$\pm$0.87 &73.31$\pm$7.33 &82.93$\pm$2.18
   &85.07$\pm$1.71 &93.54$\pm$0.52  &95.45$\pm$0.26 &\textbf{96.23$\pm$0.50} \\  \hline

 \end{tabular}
}
\label{table:Reuters}
\end{table*}

\begin{table*}[t]
\caption{Comparison results in terms of 3 different metrics (mean $\pm$ standard deviation) on 20NewsGroups dataset.}
\centering
\scalebox{0.77}{
\begin{tabular}{|cc||ccc||ccccc||c|}
\hline
 {}&Metrics&stSC & uSC & OnestepSC& MBC& SMBC &SMKC & MTSC  & MTCMRL &Ours \\
 \hline\hline
  \multirow {3}{*}{
  \centering Task1}
   & Purity($\%$) &63.89$\pm$0.15  & 44.52$\pm$0.49 &66.53$\pm$1.98  &47.69$\pm$2.13  &50.45$\pm$5.41 & 73.89$\pm$1.36 &77.27$\pm$0.78    &\textbf{81.59$\pm$1.45}   & 81.05$\pm$1.05 \\

   &NMI($\%$)& 30.77$\pm$0.33  & 4.35$\pm$0.33 &38.74$\pm$1.10     &19.29$\pm$2.76   &24.80$\pm$3.18   & 37.75$\pm$2.68  &45.35$\pm$0.83 & \textbf{49.38$\pm$1.55}  &46.38$\pm$0.62 \\

   &RI($\%$)& 61.27$\pm$0.30  & 56.32$\pm$0.24 &65.54$\pm$1.48  &48.93$\pm$7.45  &54.19$\pm$0.72  & 72.09$\pm$1.17  &74.31$\pm$0.69     &78.45$\pm$1.47 & \textbf{78.60$\pm$0.15} \\ \hline
   \multirow {3}{*}{
  \centering Task2}
    &  Purity($\%$) & 53.54$\pm$0.48  & 40.89$\pm$0.00 & 55.97$\pm$0.13  &48.56$\pm$2.96  &50.46$\pm$1.31  &66.81$\pm$1.44 &63.55$\pm$0.78 & 65.06$\pm$0.77&\textbf{73.47$\pm$0.09}  \\

    & NMI($\%$) &  34.68$\pm$0.20  & 9.92$\pm$0.00 & 32.86$\pm$0.08 & 21.27$\pm$3.45  &23.23$\pm$7.97  &40.76$\pm$2.88 &42.52$\pm$0.33 & 44.21$\pm$0.39 &\textbf{52.75$\pm$0.41}  \\

    & RI($\%$) & 60.08$\pm$0.66  & 65.51$\pm$0.00  & 62.54$\pm$0.17  &64.31$\pm$2.16  &63.82$\pm$4.60 &76.26$\pm$1.01 &70.23$\pm$0.21 & 72.19$\pm$0.18 &\textbf{81.17$\pm$0.05}  \\   \hline
     \multirow {3}{*}{
  \centering Task3}
   & Purity($\%$) & 59.07$\pm$0.00 & 54.74$\pm$0.00  &59.87$\pm$1.68 &49.85$\pm$3.05 & 52.34$\pm$1.43 &60.40$\pm$2.15  &68.86$\pm$1.26 &77.86$\pm$0.69 &\textbf{83.73$\pm$0.11} \\

   & NMI($\%$) & 34.58$\pm$0.09  &17.63$\pm$0.00  &39.25$\pm$1.93 & 20.53$\pm$5.41 & 23.37$\pm$4.01 &30.24$\pm$1.12 &38.81$\pm$1.56 &46.05$\pm$1.31 &\textbf{55.54$\pm$0.37} \\

   & RI($\%$) & 61.08$\pm$0.01  &58.10$\pm$0.00   &61.47$\pm$1.51  &48.35$\pm$2.76 &52.67$\pm$0.89 &65.23$\pm$0.98  &64.06$\pm$1.30 &75.14$\pm$0.58 &\textbf{82.06$\pm$0.14} \\  \hline
  \multirow {3}{*}{
  \centering Task4}
   & Purity($\%$) &  51.51$\pm$0.14 & 52.35$\pm$0.45 & 54.37$\pm$0.29  &46.33$\pm$2.86 & \textbf{75.18$\pm$4.77} &68.69$\pm$0.35  & 67.35$\pm$0.35 & 74.85$\pm$0.89 &72.08$\pm$3.19 \\

   & NMI($\%$) &  32.53$\pm$0.32 & 26.13$\pm$0.87 & 34.12$\pm$0.73  &21.37$\pm$3.48 &44.09$\pm$4.78 & 41.15$\pm$0.95  & 44.03$\pm$0.31 & 54.02$\pm$0.65 & \textbf{56.71$\pm$1.33} \\

   & RI($\%$) &  52.54$\pm$0.19 & 64.70$\pm$0.25  & 56.27$\pm$0.38   &46.61$\pm$2.70 &78.99$\pm$2.71 &74.68$\pm$0.41  & 70.35$\pm$0.41 & 78.56$\pm$0.74 & \textbf{82.29$\pm$1.25} \\ \hline \hline
 \multirow {3}{*}{
  \centering }
   & Avg.Purity($\%$) &56.99$\pm$0.19 &48.12$\pm$0.23  &59.18$\pm$1.02 &48.11$\pm$2.75 & 57.01$\pm$4.35 &67.45$\pm$1.33  &69.25$\pm$0.64 &74.91$\pm$0.93 &\textbf{77.73$\pm$1.11} \\

   &Avg.NMI($\%$) & 33.03$\pm$0.24  &14.51$\pm$0.30  &36.24$\pm$0.96 &20.62$\pm$3.78 &28.12$\pm$4.98 &37.48$\pm$1.93 &42.68$\pm$0.76 &48.39$\pm$0.98 &\textbf{52.84$\pm$0.68} \\

   &Avg.RI($\%$) & 58.73$\pm$0.29  &61.16$\pm$0.12   &61.46$\pm$0.89 &52.05$\pm$3.77 &62.42$\pm$2.23 &72.07$\pm$0.89  &69.74$\pm$0.41 &76.15$\pm$0.85 &\textbf{81.12$\pm$0.39} \\  \hline

 \end{tabular}
}
\label{table:20NewsGroups}
\end{table*}

\begin{table*}[htbp]
\caption{Runtime (seconds) on a standard CPU of all competing models.}
\centering
\scalebox{0.77}{
\begin{tabular}{|p{74pt}||ccc||ccccc||c|}
\hline
 {}&stSC & uSC & OnestepSC& MBC& SMBC &SMKC & MTSC  & MTCMRL &Ours \\
 \hline\hline
WebKB4(s)       & 1.22$\pm$0.01  &1.21$\pm$0.03      &600.91$\pm$26.60              &6.97$\pm$1.08  &5.77$\pm$0.14 & 34.79$\pm$0.47   & 69.72$\pm$1.26    &14.51$\pm$1.30 &2.69$\pm$0.02 \\ \hline
Reuters(s)       & 0.87$\pm$0.20  &1.31$\pm$0.22      &1410.47$\pm$47.47              &3.91$\pm$0.19  &5.47$\pm$0.14 & 16.86$\pm$0.84   & 71.79$\pm$1.20    & 8.26$\pm$0.28 & 1.32$\pm$0.01 \\ \hline
20NewsGroups(s)      &  2.92$\pm$0.07   & 5.27$\pm$0.02     &3500.16$\pm$77.70          &19.19$\pm$1.04  &26.54$\pm$1.30  & 316.22$\pm$3.53     &44.01$\pm$3.53 &384.52$\pm$19.55 &9.95$\pm$0.29 \\   \hline
 \end{tabular}
}
\label{table:runtime}
\end{table*}


\section{Experiments}\label{sec:experiment}
This section evaluates the clustering performance of our proposed $\mr{L^2SC}$ model via several empirical comparisons. We firstly introduce the used competing models. Several adopted datasets and experimental results are then provided, followed by some analyses of our model.

\subsection{Comparison Models and Evaluation}
The experiments in this subsection evaluate our proposed $\mr{L^2SC}$ model with three single spectral clustering models, and five multi-task clustering models.

\emph{Single spectral clustering models}: 1) Spectral Clustering (stSC) \cite{ng2002spectral}: standard spectral clustering model; 2) Spectral clustering-union (uSC) \cite{ng2002spectral}: spectral clustering model, which can be achieved via collecting all the clustering task data (\emph{i.e.,} ``pooling'' all the task data and ignoring the multi-task setting); 3) One-step spectral clustering (OnestepSC) \cite{zhu2017one}: single spectral clustering task model.

\emph{Multi-task clustering models}: 1) Multi-task Bregman Clustering (MBC) \cite{zhang2011multitask}: this model consists of average Bregman divergence and a task regularization; 2) Smart Multi-task Bregman Clustering (SMBC) \cite{zhang2015smart}: unsupervised transfer learning model, which focuses on clustering a small collection of target unlabeled data with the help of auxiliary unlabeled data; 3) Smart Multi-task Kernel Clustering (SMKC)~\cite{zhang2015smart}: this model can deal with nonlinear data by introducing Mercer kernel; 4) Multi-Task Spectral Clustering (MTSC)~\cite{yang2015multitask}: this model performs spectral clustering over multiple related tasks by using their inter-task correlations; 5) Multi-Task Clustering with Model Relation Learning (MTCMRL)~\cite{zhang2018multi}: this model can automatically learn the model parameter relatedness between each pair of tasks.

For the evaluation, we adopt three performance measures: normalized mutual information (NMI), clustering purity (Purity) and rand index (RI) \cite{schutze2008introduction} to evaluate the clustering performance. The bigger the value of NMI, Purity and RI is, the better the clustering performance of the corresponding model will be. We implement all the models in MATLAB, and all the used parameters of the models are tuned in $\{10^{-3}\times i\}_{i=1}^{10}\cup \{10^{-2}\times i\}_{i=2}^{10}\cup \{10^{-1}\times i\}_{i=2}^{10}\cup \{2\times i\}_{i=1}^{10}\cup \{40\times i\}_{i=1}^{20}$. Although different $\lambda_t$'s are allowed for different tasks in our model, this paper we only differentiate between $\mu$ and $\lambda=\lambda_{t}>0 $.



\subsection{Real Datasets $\&$ Experiment Results}
According to whether the number of cluster center is consistent or not, there are two different scenarios for multi-task clustering tasks: \textbf{Cluster-consistent} and \textbf{Cluster-inconsistent}. For the \textbf{Cluster-consistent} dataset, it can be roughly divided into: same clustering task and different clustering tasks with same number of cluster centers. We thus use two datasets in this paper: WebKB4\footnote{http://www.cs.cmu.edu/afs/cs.cmu.edu/project/theo20/www \\/data/} with 2500 dimensions and Reuters\footnote{http://www.cad.zju.edu.cn/home/dengcai/Data/TextData.html} with 6370 dimensions, respectively. For the WebKB4 dataset, which includes web pages collected from computer science department websites at 4 universities: Cornell, Texas, Washington and Wisconsin, and 7 categories. Following the setting in \cite{zhang2018multi}, 4 most populous categories (\emph{i.e.,} course, faculty, project and student) are chosen for clustering. Accordingly, for the Reuters dataset, 4 most populous root categories (\emph{i.e.,} economic index, energy, food and metal) are chosen for clustering, and the total number of task is 3. For the \textbf{Cluster-inconsistent} dataset, we also adopt 20NewsGroups\footnote{http://qwone.com/~jason/20Newsgroups/} dataset with 3000 dimensions by following \cite{zhang2018multi}, which consists of the news documents under 20 categories. Since ``negative transfer'' \cite{zhou2015flexible} will happen when the cluster centers of multiple consecutive spectral tasks have significant changes, 4 most populous root categories (\emph{i.e.,} comp, rec, sci and talk) are selected for clustering, while the 1-th and 3-th tasks are set as 3 categories, and the 2-th and 4-th tasks are set as 4 categories.

The experimental results (competing models with parameter setting are averaged over 10 random repetitions) are provided in Table~\ref{table:WebKB4}, Table~\ref{table:Reuters} and Table~\ref{table:20NewsGroups}, where the task sequence for our $\mr{L^2SC}$ is in a random way. From the presented results, we can notice that: \textbf{1)} Our proposed lifelong spectral clustering model outperforms the single-task spectral clustering methods, since $\mr{L^2SC}$ can exploit the information among multiple related tasks, whereas the single-task spectral clustering model only use the information within each task. MTCMRL performs worse than our proposed $\mr{L^2SC}$ in most cases, because even though it incorporates the cross-task relatedness with the linear regression model, it does not consider the feature embedding correlations among each pair of clustering tasks. The reason why MTCMRL performs better than our $\mr{L^2SC}$ in Task1 of 20NewsGroups is that we set $k=4$ in this \textbf{Cluster-inconsistent} dataset, whereas the number of cluster center is $3$ in Task1. \textbf{2)} In addition to MTCMRL and single-task spectral clustering models, our $\mr{L^2SC}$ performs much better than the comparable multi-task clustering model cases. It is because that $\mr{L^2SC}$ can not only learn the latent cluster center between each pair of tasks via the orthogonal basis library $B$, but also control the number of embedded features common across the clustering tasks. \textbf{3)} Additionally, Table~\ref{table:runtime} also shows that the runtime comparisons between our $\mr{L^2SC}$ model and other single/multi-task clustering models. $\mr{L^2SC}$ is faster and better than the most multi-task clustering models on WebKB, Reuters and 20NewsGroups datasets, \emph{e.g.,} SMBC and MTSC, also OnestepSC. However, $\mr{L^2SC}$ is little slower than stSC and uSC. This is because both stSC and uSC can obtain the cluster assignment matrix via closed-form solution, \emph{i.e.,} eigenvalue decomposition of the $K^t$ in Eq.~\eqref{eq:spectralcluster_single}. We perform all the experiments on the computer with Intel i7 CPU, 8G RAM.

\begin{figure}[t]
\center
  \hspace{-3.1mm}
 \subcaptionbox{WebKB4 Dataset}{
  \centering
   \includegraphics[trim = 0mm 0mm 0mm 0mm, clip, width=0.23\textwidth]{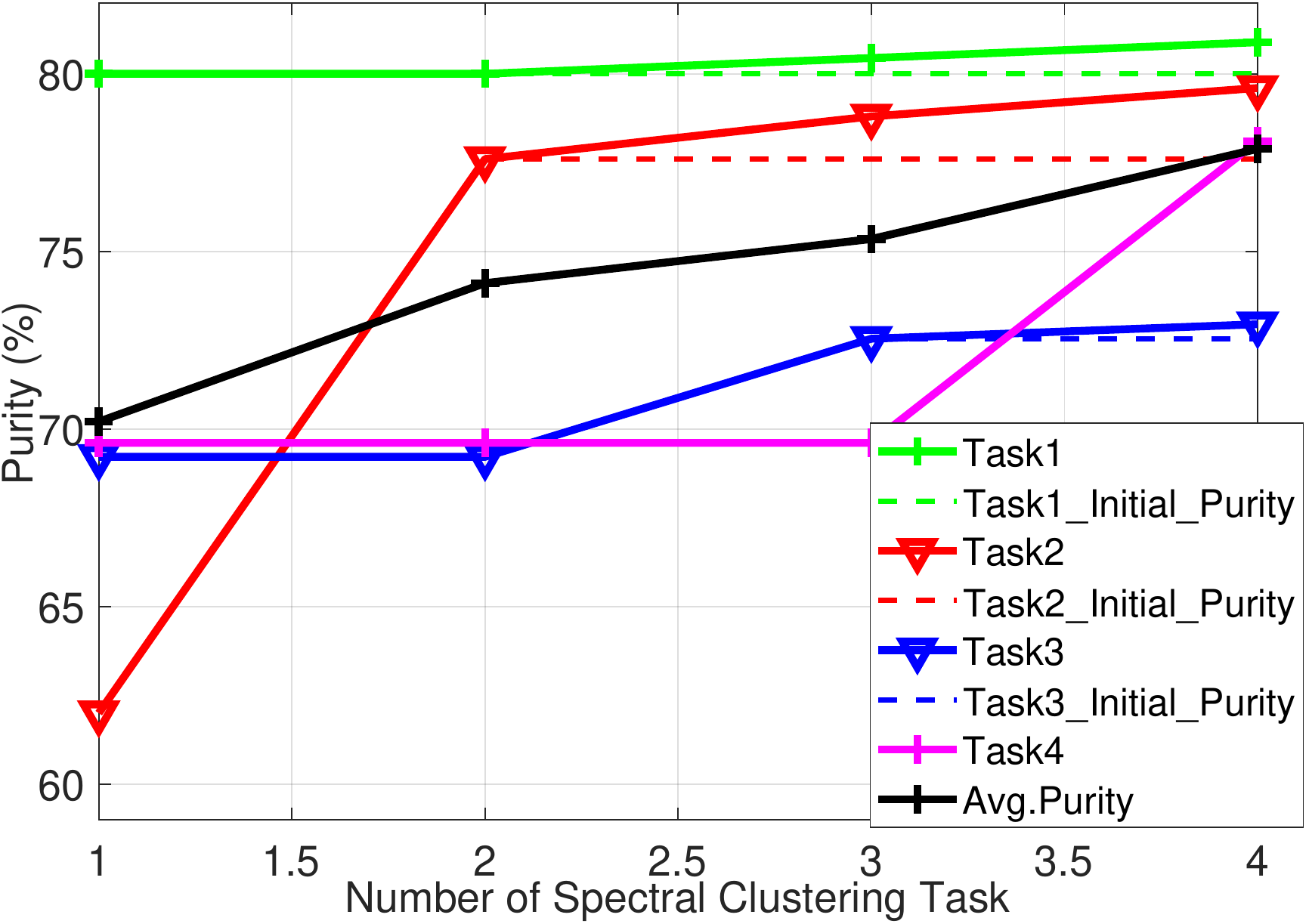}
  }
     \hspace{-1.6mm}
 \subcaptionbox{WebKB4 Dataset}{
  \centering
   \includegraphics[trim = 0mm 0mm 0mm 0mm, clip, width=0.23\textwidth]{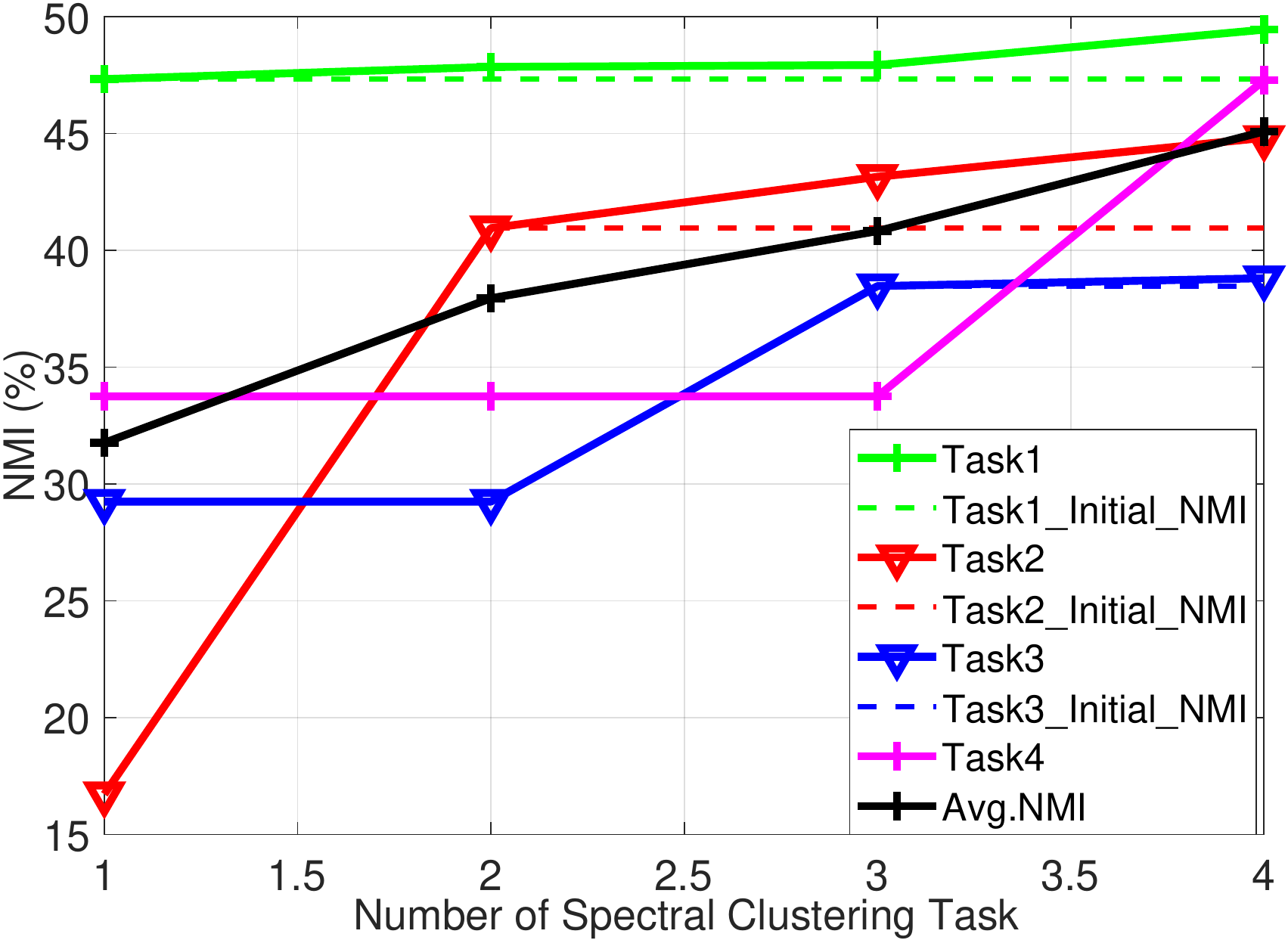}
}
     \hspace{-1.6mm}
       \caption{The influence of the number of learned tasks on WebKB4 datasets in terms of Purity and NMI metrics, where the vertical and horizontal axes denote the clustering performance and number of learned tasks, respectively. The initial clustering performance of each task (except for the first task) of each dataset is achieved using stSC algorithm.}
    \label{fig:lifelonglearning}
\end{figure}



\begin{figure}[t]
\center
  \hspace{-3.1mm}
\subcaptionbox{WebKB4 Dataset}{
  \centering
   \includegraphics[trim = 0mm 0mm 0mm 0mm, clip, width=0.23\textwidth]{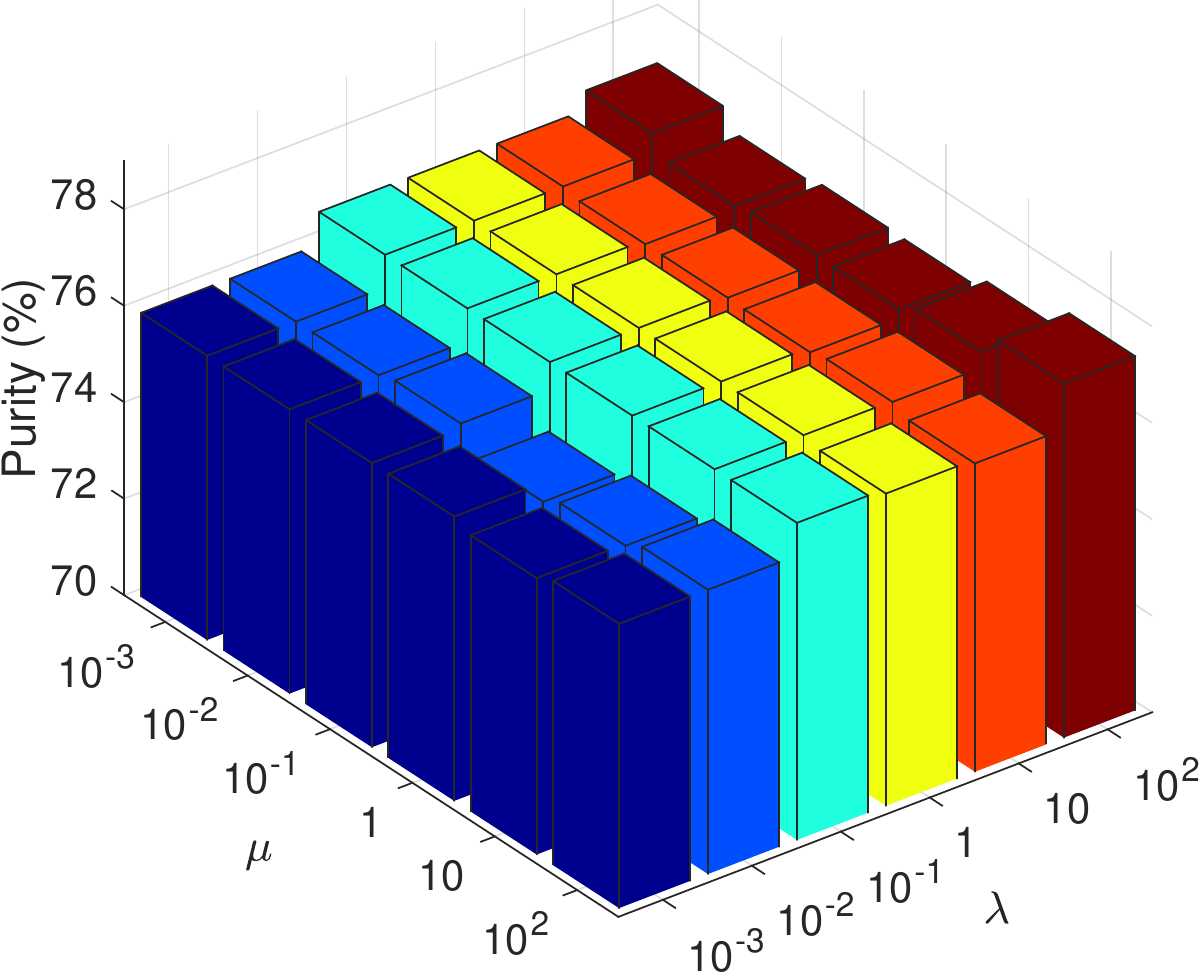}
     }
     \hspace{-1.6mm}
  \subcaptionbox{WebKB4 Dataset}{
  \centering
   \includegraphics[trim = 0mm 0mm 0mm 0mm, clip, width=0.23\textwidth]{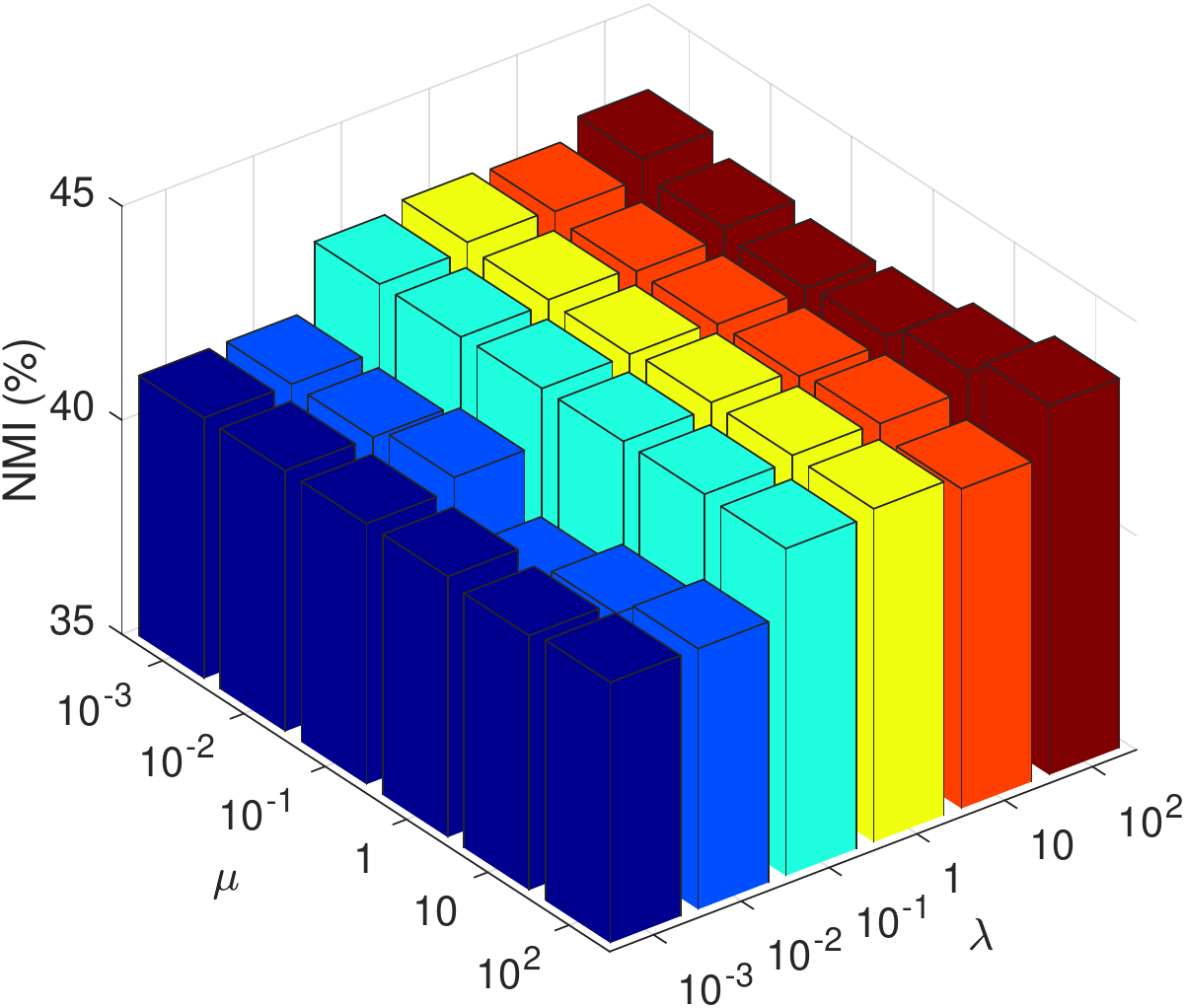}
  }
     \hspace{-1.6mm}

    \caption{Parameter analysis of our proposed $\mr{L^2SC}$ model on WebKB4 dataset.}
    \label{fig:parameter}

\end{figure}

\textbf{Evaluating Lifelong learning:} This subsection studies the lifelong learning property of our $\mr{L^2SC}$ model by following \cite{ruvolo2013ella}, \emph{i.e.,} how well the clustering performance will be as the number of clustering tasks $t$ increases. We adopt the WebKB4 dataset, set the sequence of learned $t$ tasks as: Task1, Task2, Task3 and Task4, and present the clustering performance in Figure~\ref{fig:lifelonglearning}. Obviously, as new clustering task is imposed step-by-step, the performances (\emph{i.e.,} Purity and NMI) for both learned and learning task are improved gradually when comparing with stSC (initial clustering result of each line in Figure~\ref{fig:lifelonglearning}), which justifies $\mr{L^2SC}$ can accumulate continually knowledge and accomplish lifelong learning just like “human learning”. Furthermore, the performance of early clustering tasks can improve obviously than succeeding ones, \emph{i.e.,} the early spectral clustering tasks can benefit more from the stored knowledge than later ones.

\textbf{Parameter Investigation:} In order to study how the parameters $\lambda$ and $\mu$ affect the clustering performance of our $\mr{L^2SC}$. For the WebKB4 dataset, we repeat the $\mr{L^2SC}$ ten times by fixing one parameter and tuning the other parameters in $[0.001,0.01,0.1,1,10,100]$. As depicted in Figure~\ref{fig:parameter}, we can notice that clustering performance changes with different ratio of parameters, which give the evidence that the appropriate parameters can make the generalization performance better, \emph{e.g.,} $\lambda=100$ for WebKB4 dataset.

\begin{figure}[t]
\center
  \hspace{-3.1mm}
 \subcaptionbox{WebKB4 Dataset}{
  \centering
   \includegraphics[trim = 0mm 0mm 0mm 0mm, clip, width=0.23\textwidth]{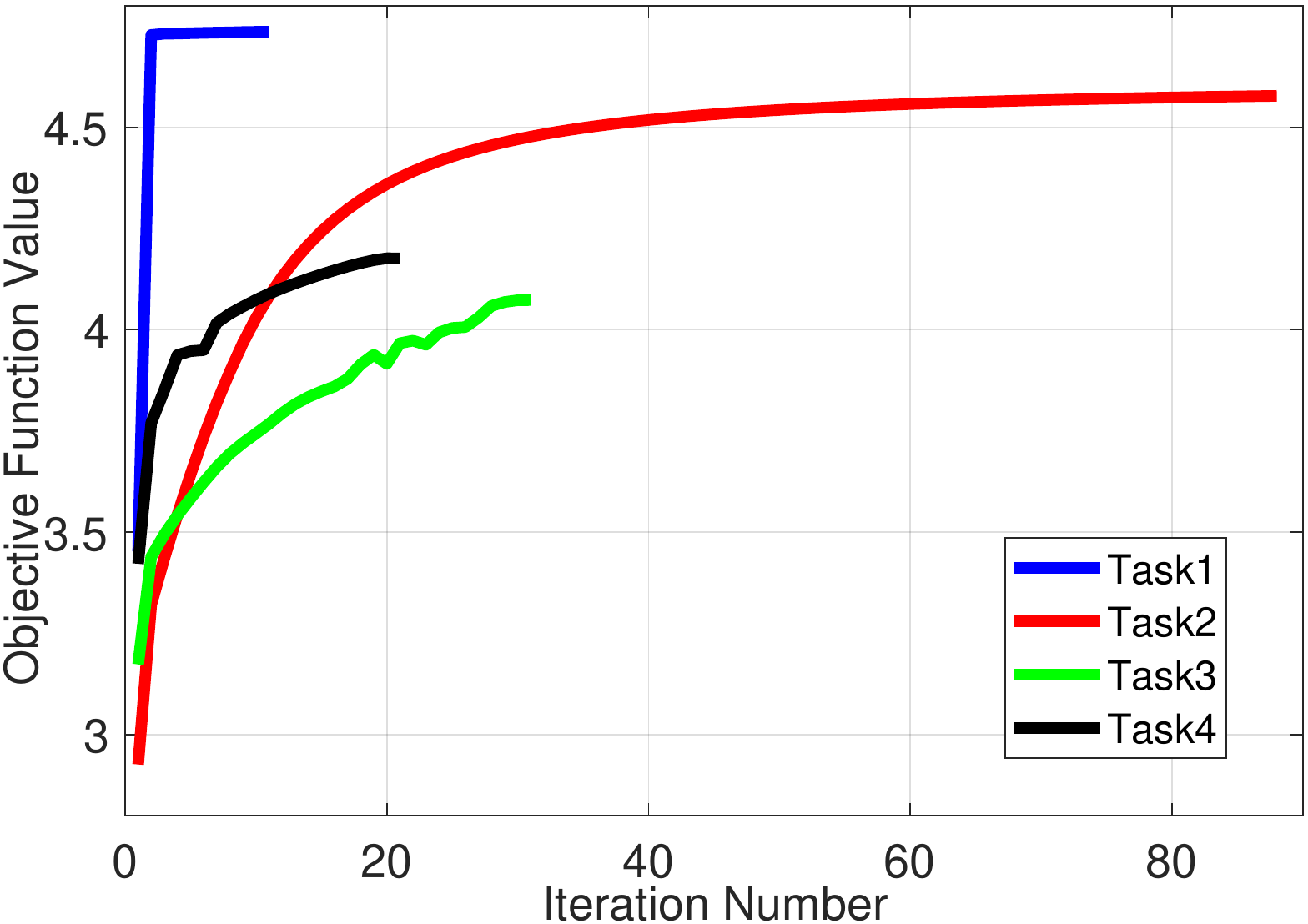}
}
 \subcaptionbox{20NewsGroups Dataset}{
  \centering
   \includegraphics[trim = 0mm 0mm 0mm 0mm, clip, width=0.23\textwidth]{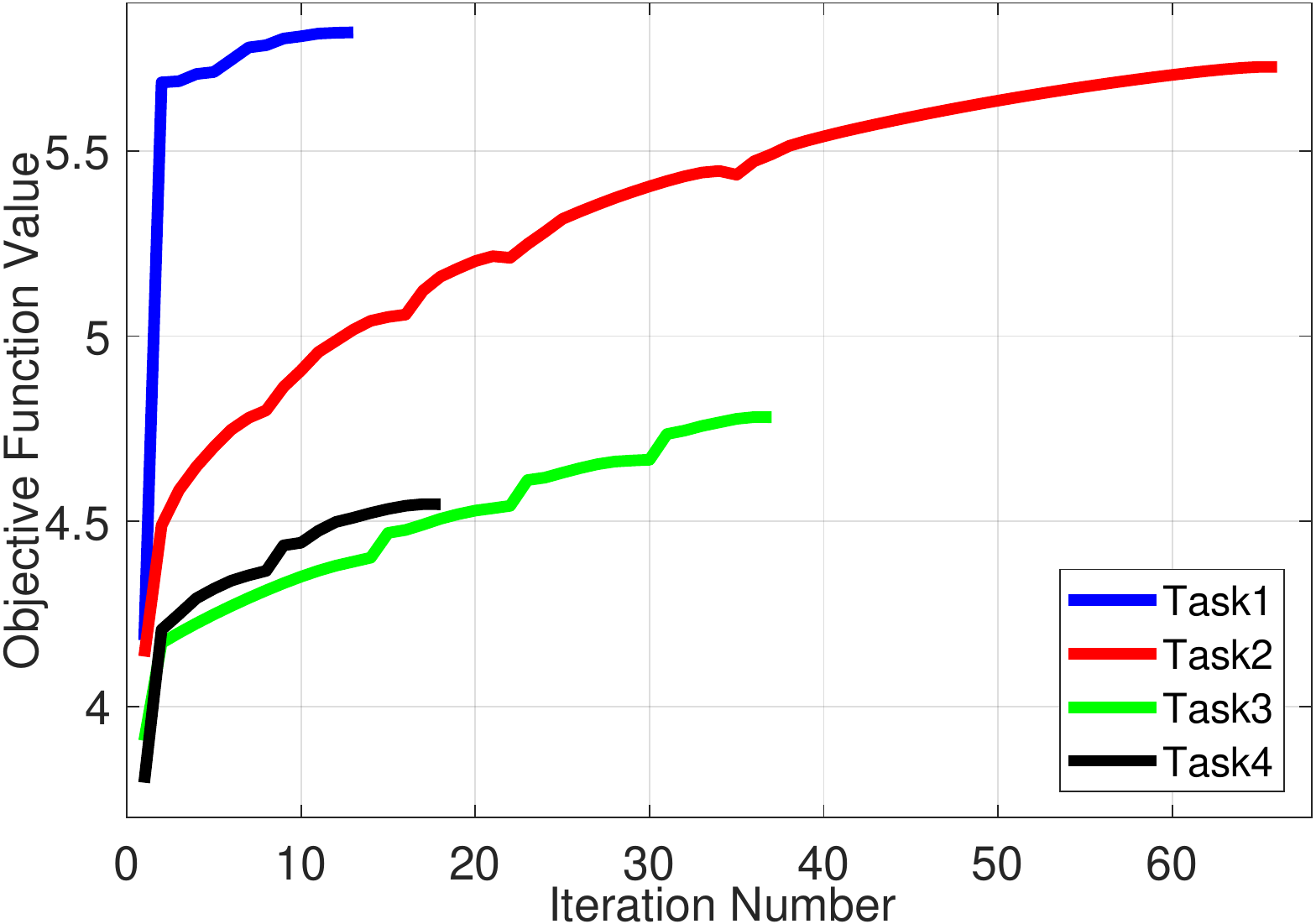}
}
     \hspace{-1.6mm}

         \caption{Convergence analysis of our proposed $\mr{L^2SC}$ model on (a) WebKB4 and (b) 20NewsGroups datasets, where lines with different colors denote different tasks in each dataset.}
    \label{fig:convergence}
\end{figure}

\textbf{Convergence Analysis:} To investigate the convergence of our proposed optimisation algorithm for solving $\mr{L^2SC}$ model, we plot the value of total loss terms for each new task on WebKB4 and 20NewsGroups datasets. As shown in Figure~\ref{fig:convergence}, the objective function values increase with respect to iterations, and the values for each new task approach to be a fixed point after a few iterations (\emph{e.g.,} less than 20 iteration for Task 4 on both datasets), \emph{i.e.,} although the convergence analysis of $\mr{L^2SC}$ cannot be proved directly in our paper, we find it converge asymptotically on the real-world datasets.


\section{Conclusion} \label{sec:conclusion}
This paper studies how to add spectral clustering capability into original spectral clustering system without damaging existing capabilities. Specifically, we propose a lifelong learning model by incorporating spectral clustering: lifelong spectral clustering ($\mr{L^2SC}$), which learns a library of orthogonal basis as a set of latent cluster centers, and a library of embedded features for all the spectral clustering tasks. When a new spectral clustering task arrives, $\mr{L^2SC}$ can transfer knowledge embedded in the shared knowledge libraries to encode the coming spectral clustering task with encoding matrix, and redefine the libraries with the fresh knowledge. We have conducted experiments on several real-world datasets; the experimental results demonstrate the effectiveness and efficiency of our proposed $\mr{L^2SC}$ model.



\bibliographystyle{aaai}
\bibliography{LifelongSpectralClustering}

\begin{thebibliography}{}

\bibitem[\protect\citeauthoryear{Ammar \bgroup et al\mbox.\egroup
  }{2014}]{ammar2014online}
Ammar, H.~B.; Eaton, E.; Ruvolo, P.; and Taylor, M.
\newblock 2014.
\newblock Online multi-task learning for policy gradient methods.
\newblock In {\em ICML-14},  1206--1214.

\bibitem[\protect\citeauthoryear{Argyriou, Evgeniou, and
  Pontil}{2008}]{Argyriou:2008}
Argyriou, A.; Evgeniou, T.; and Pontil, M.
\newblock 2008.
\newblock Convex multi-task feature learning.
\newblock {\em Machine Learning} (73):243--272.

\bibitem[\protect\citeauthoryear{Chen, Ma, and Liu}{2018}]{chen2018lifelong}
Chen, Z.; Ma, N.; and Liu, B.
\newblock 2018.
\newblock Lifelong learning for sentiment classification.
\newblock {\em arXiv preprint arXiv:1801.02808}.

\bibitem[\protect\citeauthoryear{Han and Kim}{2015}]{han2015unsupervised}
Han, D., and Kim, J.
\newblock 2015.
\newblock Unsupervised simultaneous orthogonal basis clustering feature
  selection.
\newblock In {\em CVPR},  5016--5023.

\bibitem[\protect\citeauthoryear{Hinton, Vinyals, and
  Dean}{2015}]{hinton2015distilling}
Hinton, G.; Vinyals, O.; and Dean, J.
\newblock 2015.
\newblock Distilling the knowledge in a neural network.
\newblock {\em arXiv preprint arXiv:1503.02531}.

\bibitem[\protect\citeauthoryear{Huy \bgroup et al\mbox.\egroup
  }{2013}]{huy2013feature}
Huy, T.~N.; Shao, H.; Tong, B.; and Suzuki, E.
\newblock 2013.
\newblock A feature-free and parameter-light multi-task clustering framework.
\newblock {\em Knowledge and information systems} 36(1):251--276.

\bibitem[\protect\citeauthoryear{Isele and Cosgun}{2018}]{isele2018selective}
Isele, D., and Cosgun, A.
\newblock 2018.
\newblock Selective experience replay for lifelong learning.
\newblock {\em arXiv preprint arXiv:1802.10269}.

\bibitem[\protect\citeauthoryear{Isele, Rostami, and
  Eaton}{2016}]{isele2016using}
Isele, D.; Rostami, M.; and Eaton, E.
\newblock 2016.
\newblock Using task features for zero-shot knowledge transfer in lifelong
  learning.
\newblock In {\em IJCAI},  1620--1626.

\bibitem[\protect\citeauthoryear{Jiang and Chung}{2012}]{jiang2012transfer}
Jiang, W., and Chung, F.-l.
\newblock 2012.
\newblock Transfer spectral clustering.
\newblock In {\em Joint European Conference on Machine Learning and Knowledge
  Discovery in Databases},  789--803.
\newblock Springer.

\bibitem[\protect\citeauthoryear{Kang \bgroup et al\mbox.\egroup
  }{2018}]{kang2018unified}
Kang, Z.; Peng, C.; Cheng, Q.; and Xu, Z.
\newblock 2018.
\newblock Unified spectral clustering with optimal graph.
\newblock In {\em AAAI}.

\bibitem[\protect\citeauthoryear{Li and Chen}{2015}]{li2015superpixel}
Li, Z., and Chen, J.
\newblock 2015.
\newblock Superpixel segmentation using linear spectral clustering.
\newblock In {\em CVPR},  1356--1363.

\bibitem[\protect\citeauthoryear{Li and Hoiem}{2016}]{li2016learning}
Li, Z., and Hoiem, D.
\newblock 2016.
\newblock Learning without forgetting.
\newblock In {\em ECCV},  614--629.

\bibitem[\protect\citeauthoryear{Manton}{2002}]{manton2002optimization}
Manton, J.~H.
\newblock 2002.
\newblock Optimization algorithms exploiting unitary constraints.
\newblock {\em IEEE Transactions on Signal Processing} 50(3):635--650.

\bibitem[\protect\citeauthoryear{Ng, Jordan, and Weiss}{2002}]{ng2002spectral}
Ng, A.~Y.; Jordan, M.~I.; and Weiss, Y.
\newblock 2002.
\newblock On spectral clustering: Analysis and an algorithm.
\newblock In {\em Advances in neural information processing systems},
  849--856.

\bibitem[\protect\citeauthoryear{Nie \bgroup et al\mbox.\egroup
  }{2010}]{nie2010efficient}
Nie, F.; Huang, H.; Cai, X.; and Ding, C.~H.
\newblock 2010.
\newblock Efficient and robust feature selection via joint l2, 1-norms
  minimization.
\newblock In {\em NIPS},  1813--1821.

\bibitem[\protect\citeauthoryear{Pang \bgroup et al\mbox.\egroup
  }{2018}]{pang2018spectral}
Pang, Y.; Xie, J.; Nie, F.; and Li, X.
\newblock 2018.
\newblock Spectral clustering by joint spectral embedding and spectral
  rotation.
\newblock {\em IEEE transactions on cybernetics}.

\bibitem[\protect\citeauthoryear{Pentina and
  Lampert}{2015}]{pentina2015lifelong}
Pentina, A., and Lampert, C.~H.
\newblock 2015.
\newblock Lifelong learning with non-iid tasks.
\newblock In {\em Advances in Neural Information Processing Systems},
  1540--1548.

\bibitem[\protect\citeauthoryear{Rannen Ep~Triki \bgroup et al\mbox.\egroup
  }{2017}]{rannen2017encoder}
Rannen Ep~Triki, A.; Aljundi, R.; Blaschko, M.; and Tuytelaars, T.
\newblock 2017.
\newblock Encoder based lifelong learning.
\newblock In {\em Proceedings ICCV 2017},  1320--1328.

\bibitem[\protect\citeauthoryear{Ruvolo and Eaton}{2013}]{ruvolo2013ella}
Ruvolo, P., and Eaton, E.
\newblock 2013.
\newblock Ella: An efficient lifelong learning algorithm.
\newblock In {\em ICML},  507--515.

\bibitem[\protect\citeauthoryear{Sch{\"u}tze, Manning, and
  Raghavan}{2008}]{schutze2008introduction}
Sch{\"u}tze, H.; Manning, C.~D.; and Raghavan, P.
\newblock 2008.
\newblock {\em Introduction to information retrieval}, volume~39.
\newblock Cambridge University Press.

\bibitem[\protect\citeauthoryear{Shi and Malik}{2000}]{shi2000normalized}
Shi, J., and Malik, J.
\newblock 2000.
\newblock Normalized cuts and image segmentation.
\newblock {\em IEEE Transactions on pattern analysis and machine intelligence}
  22(8):888--905.

\bibitem[\protect\citeauthoryear{Sun \bgroup et al\mbox.\egroup
  }{2017}]{sun2017joint}
Sun, G.; Cong, Y.; Hou, D.; Fan, H.; Xu, X.; and Yu, H.
\newblock 2017.
\newblock Joint household characteristic prediction via smart meter data.
\newblock {\em IEEE Transactions on Smart Grid}.

\bibitem[\protect\citeauthoryear{Sun \bgroup et al\mbox.\egroup
  }{2018a}]{sun2018robust}
Sun, G.; Cong, Y.; Li, J.; and Fu, Y.
\newblock 2018a.
\newblock Robust lifelong multi-task multi-view representation learning.
\newblock In {\em 2018 IEEE International Conference on Big Knowledge (ICBK)},
  91--98.
\newblock IEEE.

\bibitem[\protect\citeauthoryear{Sun \bgroup et al\mbox.\egroup
  }{2018b}]{sun2018lifelong}
Sun, G.; Cong, Y.; Liu, J.; Liu, L.; Xu, X.; and Yu, H.
\newblock 2018b.
\newblock Lifelong metric learning.
\newblock {\em IEEE transactions on cybernetics} (99):1--12.

\bibitem[\protect\citeauthoryear{Sun \bgroup et al\mbox.\egroup
  }{2019}]{sun2019representative}
Sun, G.; Cong, Y.; Wang, Q.; Zhong, B.; and Fu, Y.
\newblock 2019.
\newblock Representative task self-selection for flexible clustered lifelong
  learning.
\newblock {\em arXiv preprint arXiv:1903.02173}.

\bibitem[\protect\citeauthoryear{Sun, Cong, and Xu}{2018}]{sun2018active}
Sun, G.; Cong, Y.; and Xu, X.
\newblock 2018.
\newblock Active lifelong learning with" watchdog".
\newblock In {\em AAAI}.

\bibitem[\protect\citeauthoryear{Thrun and
  O'Sullivan}{1996}]{thrun1996discovering}
Thrun, S., and O'Sullivan, J.
\newblock 1996.
\newblock Discovering structure in multiple learning tasks: The tc algorithm.
\newblock In {\em ICML},  489--497.

\bibitem[\protect\citeauthoryear{Thrun}{2012}]{thrun2012explanation}
Thrun, S.
\newblock 2012.
\newblock {\em Explanation-based neural network learning: A lifelong learning
  approach}, volume 357.
\newblock Springer Science \& Business Media.

\bibitem[\protect\citeauthoryear{Wang, Ding, and Fu}{2019}]{Seg_Lichen_TIP18}
Wang, L.; Ding, Z.; and Fu, Y.
\newblock 2019.
\newblock Low-rank transfer human motion segmentation.
\newblock {\em IEEE Transactions on Image Processing} 28(2):1023--1034.

\bibitem[\protect\citeauthoryear{Xu \bgroup et al\mbox.\egroup
  }{2018}]{xu2018lifelong}
Xu, H.; Liu, B.; Shu, L.; and Yu, P.~S.
\newblock 2018.
\newblock Lifelong domain word embedding via meta-learning.
\newblock {\em arXiv preprint arXiv:1805.09991}.

\bibitem[\protect\citeauthoryear{Yang \bgroup et al\mbox.\egroup
  }{2015}]{yang2015multitask}
Yang, Y.; Ma, Z.; Yang, Y.; Nie, F.; and Shen, H.~T.
\newblock 2015.
\newblock Multitask spectral clustering by exploring intertask correlation.
\newblock {\em IEEE transactions on cybernetics} 45(5):1083--1094.

\bibitem[\protect\citeauthoryear{Zhang and Zhang}{2011}]{zhang2011multitask}
Zhang, J., and Zhang, C.
\newblock 2011.
\newblock Multitask bregman clustering.
\newblock {\em Neurocomputing} 74(10):1720--1734.

\bibitem[\protect\citeauthoryear{Zhang \bgroup et al\mbox.\egroup
  }{2017}]{zhang2017multi}
Zhang, X.; Zhang, X.; Liu, H.; and Liu, X.
\newblock 2017.
\newblock Multi-task clustering through instances transfer.
\newblock {\em Neurocomputing} 251:145--155.

\bibitem[\protect\citeauthoryear{Zhang \bgroup et al\mbox.\egroup
  }{2018}]{zhang2018multi}
Zhang, X.; Zhang, X.; Liu, H.; and Luo, J.
\newblock 2018.
\newblock Multi-task clustering with model relation learning.
\newblock In {\em IJCAI},  3132--3140.

\bibitem[\protect\citeauthoryear{Zhang, Zhang, and Liu}{2015}]{zhang2015smart}
Zhang, X.; Zhang, X.; and Liu, H.
\newblock 2015.
\newblock Smart multitask bregman clustering and multitask kernel clustering.
\newblock {\em ACM Transactions on Knowledge Discovery from Data (TKDD)}
  10(1):8.

\bibitem[\protect\citeauthoryear{Zhao, Ding, and Fu}{2017}]{zhao2017multi}
Zhao, H.; Ding, Z.; and Fu, Y.
\newblock 2017.
\newblock Multi-view clustering via deep matrix factorization.
\newblock In {\em AAAI},  2921--2927.
\newblock AAAI Press.

\bibitem[\protect\citeauthoryear{Zhou and Zhao}{2015}]{zhou2015flexible}
Zhou, Q., and Zhao, Q.
\newblock 2015.
\newblock Flexible clustered multi-task learning by learning representative
  tasks.
\newblock {\em IEEE transactions on pattern analysis and machine intelligence}
  38(2):266--278.

\bibitem[\protect\citeauthoryear{Zhu \bgroup et al\mbox.\egroup
  }{2017}]{zhu2017one}
Zhu, X.; He, W.; Li, Y.; Yang, Y.; Zhang, S.; Hu, R.; and Zhu, Y.
\newblock 2017.
\newblock One-step spectral clustering via dynamically learning affinity matrix
  and subspace.
\newblock In {\em AAAI},  2963--2969.

\end{thebibliography}

\end{document}